\documentclass{article}
\usepackage{enumerate}
\usepackage{multirow}
\usepackage[bottom,flushmargin,hang,multiple]{footmisc}
\usepackage{etoolbox}
\makeatletter
\patchcmd\maketitle{\hb@xt@1.8em}{\hbox}{}{}
\makeatother
\patchcmd{\maketitle}{\@fnsymbol}{\@arabic}{}{}
\usepackage{amsmath}
\usepackage{graphicx}
\usepackage{caption}
\usepackage{subcaption}
\usepackage[margin=1.5in]{geometry}
\usepackage{float}
\usepackage{comment}
\usepackage{pgf}
\usepackage{hyperref}

\usepackage{tikz}
\usepackage{subcaption}

\usepackage[T1]{fontenc}
\usepackage[sfdefault,scaled=.95,light]{FiraSans}
\usepackage{newtxsf}

\usepackage{etoolbox}
\AtBeginEnvironment{quote}{\small}

\begin{document}

\title{Introducing a new high-resolution handwritten digits data set with writer characteristics}



\author{C\'edric Beaulac \and Jeffrey S. Rosenthal}



\date{\today}

\maketitle

\begin{abstract}
The contributions in this article are two-fold. First, we introduce a new handwritten digit data set that we collected. It contains high-resolution images of handwritten digits together with various writer characteristics which are not available in the well-known MNIST database. The multiple writer characteristics gathered are a novelty of our data set and create new research opportunities. The data set is publicly available online.  Second, we analyse this new data set. We begin with simple supervised tasks. We assess the predictability of the writer characteristics gathered, the effect of using some of those characteristics as predictors in classification task and the effect of higher resolution images on classification accuracy. We also explore semi-supervised applications; we can leverage the high quantity of handwritten digits data sets already existing online to improve the accuracy of various classifications task with noticeable success. Finally, we also demonstrate the generative perspective offered by this new data set; we are able to generate images that mimics the writing style of specific writers. The data set has unique and distinct features and our analysis establishes benchmarks and showcases some of the new opportunities made possible with this new data set.
\end{abstract}

\pagebreak

\section{Introduction}

Modern computer vision algorithms have become impressively good at identifying the content of a complex image. A scanned handwritten document is an example of a complex image for which many algorithms were developed. In this case, the task assigned to the algorithm is to identify letters, digits and later words and sentences. In handwritten document analysis, the MNIST data set introduced by LeCun \& al. \cite{Lecun98} quickly became a benchmark for handwritten digits recognition and is now a rite of passage for computer vision algorithms.  

\bigskip

Usually, MNIST is used for a simple task, try to identify the digit in new handwritten digit images given a training set of labelled handwritten digit images. In this classification problem, the thickness of the stroke, the orientation of the digit and the general handwriting style are noise factors we are trying to overlook in order to better classify the digit itself. In this project we are interested in those handwriting style-related variables and want to understand if we can leverage those to identify the writer and various characteristics such as handedness. We attempt those prediction tasks on a brand new data set that we collected precisely for the purpose of this experiment. It is a unique data set that contains a wide range of writer characteristics unavailable in any other handwritten digits data set. 

\bigskip

We tackle the well-established task of writer identification, but also statistical inference of those writer characteristics.  Typically, computer vision algorithms are built to identify the content of the images but the tasks we tackle here are slightly more complex as we hope to predict writer characteristics that should affect only subtle details of the image. We want to explore if computer vision algorithms can capture the handwritten style and use it for inference and for image synthesis. Our contribution is two-fold; first, we introduce and distribute a new data set that we collected, HWD+, containing handwritten digit images in high-resolution and various writer characteristics. This data set can be utilized as a standalone data set and also in conjuncture with MNIST, or other handwritten digit data sets, for semi-supervised learning projects. Second, we perform a first analysis of the data set under both the supervised and semi-supervised paradigm. The supervised analysis establishes the existence of signals between images and some writer characteristics while the semi-supervised analysis demonstrate how to utilize larger data base to improve the classification performance of computer vision algorithms on the new tasks made possible by our new data set. We also showcase how computer vision algorithms are able to capture the stroke style of a writer and mimic that style in conditional image generation applications. 

\bigskip

The remaining of this paper is organized as follows: we discuss the related publications in Section 2. Section 3 introduce the new data set we collected and include some descriptive statistics.  Finally, Section 4 contains a short analysis of the HWD+ data set that highlights some of the research opportunities this data can generate. 

\section{Related Work}\label{litrev}

\subsection{Classification on MNIST}

In the contributed data set, we collected various characteristics about our writers and also assigned a writer ID to each writer. Consequently, one natural problem to tackle is writer identification. This problem has been extensively studied in the past and is still a relevant problem in forensics. A recent publication from Adak et al. \cite{Adak19} attempts to solve a writer identification problem and compares the performances of models that rely on hand-crafted feature against models with auto-derived features. Slightly before that, Xiong et al. \cite{Xiong17} produced one of the most recent surveys comparing various modern writer identification algorithms. A result shared across both articles \cite{Xiong17,Adak19} and highlighted in a comprehensive review \cite{Rehman19-2} is that auto-derived feature models perform better than feature engineering and thus we rely on auto-derived feature models in this analysis.

\bigskip

One of the tasks we have established for this research project is to assess the abilities of modern computer vision algorithms to infer some of the writer's characteristics. Most literature that discusses writer characteristics addresses graphology; the analysis of hand-writing patterns in order to identify psychological traits of writers. However, serious studies on graphology demonstrate that it is more a pseudo-science than anything else \cite{Klimoski83}. As a result, we focus this works on measurable characteristics such as age, gender or native language. We are interested in determining the feasibility of predicting such characteristics based on handwritten digits. 

\bigskip

When considering the identification of the digits themselves, the MNIST data set inspired a gigantic amount of publications. The first article to discuss this data set \cite{Lecun98} was published in 1998 and introduced the data set and compared the prediction accuracy of multiple classification methods. The two best-performing algorithms was a committee (ensemble) of deep convolutional neural networks (CNN) and a support vector machine (SVM) with test error rates as low as 0.7\% and 0.8\% respectively. This article really set the tone for future computer vision publications by establishing the sheer dominance both in terms of accuracy and memory requirements of CNNs. It was a pivotal point into explaining and empirically proving the benefits of automated feature extraction. It has also established the MNIST data as an important benchmark data set.

\bigskip

Since then the best results obtained from a SVM algorithm was obtained in 2002 \cite{Decoste02} with a 0.56\% error rate. Simple techniques that require no training, such as k-nearest neighbors (kNN), have achieved higher accuracy (0.54\% error rate) by allowing the algorithm to search into a set of distorted images \cite{keysers07}. The lowest error rate (0.35\%) achieved by a single NN was reported in 2010 \cite{Ciresan10}. Finally, in 2012 a committee of 35 CNNs achieved a 0.23\% test error rate \cite{Ciresan12}. A problem with MNIST is that current algorithms achieve a classification accuracy that is so high that it leaves room only for marginal improvements. The true usefulness of these improvements is hard to evaluate \cite{Hand06} as it might be caused to details that are specific to the MNIST data set and thus are not real improvement applicable to new problems. In other words, it is possible that MNIST has been overused and that some new models are \textit{overfitting} this data set. 

\subsection{Handwritten data sets}

Let us now introduce some related data sets. We have already introduced the MNSIT data set \cite{Lecun89}. It contains 70,000 images of handwritten digits, they have been centralized and size-normalized in a 28 by 28 pixels format in shades-of-grey. The only label attached to the images is the digit itself. The MNIST date set has multiple \textit{sibling data sets} such as Morpho-MNIST data set \cite{castro2019} which artificially extend MNIST by perturbing the data in meaningful ways such as thinning, thickening and local swelling of the digits. The authors built this data set to assess and diagnostics representation learning models. This data set shares similarities with ours with their focus being those subtle details and not so much the digit within the image. However, the difference lies in what we hope to do with those subtle handwritten style details; we hope to identify writers and predict some of their characteristics. 

\bigskip

Another similar data set is the USPS data set \cite{hull1994}. It includes a bit more than 9,000 images of handwritten digits that appeared in ZIP codes scanned from mail at the main post office in Buffalo, NY. Again in this data set very little is known about the writers. 

\bigskip

The ARDIS data set \cite{kusetogullari2020} images were collected from church documents written by different priests from 1895 to 1970. It is divided in four data sets ranging from \textit{Dataset I} being the least processed, containing strings of digits (years) in a 175 by 95 pixels format stored in RGB color to \textit{Dataset IV} being the most processed where digits have been separated and the images are now in shades-of-grey and size-normalized to be 28 by 28 pixels to mimic the MNIST structure. This allows researchers to combine this data set with MNIST, which is pretty similar to what we are doing as well. However, the focus of this data set is to offer a more complicated digit classification task; the paper texture, the type of ink used and the wider range of handwriting style is making this classification task more complicated and this an important aspect of this data set. The authors also offer a larger counterpart of this data set, the DIGITNET data set \cite{kusetogullari2020b}. However, once again, here the main task is the classification of the digit and strokes thickness and ink types are added noise we want to go beyond to better classify the content of the image. Again, the main difference between this data set and ours are the labels attached to each images allowing for a wide range of classification task that do not simply focus on the recognition of the digit. 

\bigskip

There exist multiple other handwritten data sets online \cite{indermuhle10,zhou10,santosh10,mouchere11}, in fact, there are too many to list them all. Unfortunately, none of the data set we found contain any information about the writers. 

\bigskip

There exist a few text-written data sets allowing for writer identification and some articles \cite{cha00,marti2001,srihari02} discuss this particular classification task. However, these data sets usually contain full words or complete text in contrast with the what we are doing in this current manuscript which is to predict writers simply using digits. We are interested to find out if it is indeed possible to recognize the writer given only a small amount of information; the handwriting style of 10 digits. In this aspect, the proposed tasks share similarities with signature forgery research \cite{hilton1952,ferrer2012a,ferrer2012} where the task is not about identifying the content of the images, the name contained in the signature, but rather focus on subtle details related to handwriting style to determine if the signature is a forgery and even try to identify the identity of the signature forger.

\bigskip

Compared to already existing data sets, our proposed data set is not the largest nor the most realistic set up for digit recognition. However, it contains labels such as a writer ID and writer characteristic such as handedness that were not collected in any aforementioned data sets. The controlled environment we collected the data does not create the most realistic digit recognition scenario but it allowed us to isolate the handwriting style from the pen type, paper type and other sources of variability. It also gave us replications. Because we have 14 replications for every digit/writer pair, we have enough data to capture the handwriting style and explore its connection to various writer characteristics. The replications also enabled us to explore style-conditioned handwritten digits image generation. Additionally, its simplicity, the fact that it contains strictly digits, makes this data set very accessible. 

\section{HWD+ : a new handwritten digits data set}

We named our HandWritten Digits data set HWD+, the plus sign stands for the additional writer characteristics collected. To collect a valuable data set, we followed some recommendations included in a recent article published by Rehman et al. \cite{Rehman19}. The same authors noted in another article \cite{Rehman19-2} how few existing data sets have a large enough data size to utilize modern computer vision architecture for writer identification; our data set is a contribution in that aspect.

\bigskip

The HWD+ data set contains 13,580 images from 97 different writers. Images were collected in a high resolution of 500 by 500 pixels in a shades-of-grey format. We also collected various information about the writers. We believe our data set has a weak signal for some variables and thus leave plenty of room for improvement in contrast to the popular MNIST data set where almost all algorithms achieve a good performance and where top-of-the-line algorithms achieve such a high accuracy that it becomes difficult to distinguish their performances. 

\bigskip 

We believe that the high resolution and the set of writer characteristics collected will lead to new questions and findings. It is a unique data set that could be used in multiple fashions; this is why this section carefully explains how to the data was gathered and processed into the data set now publicly available \href{https://drive.google.com/drive/folders/1f2o1kjXLvcxRgtmMMuDkA2PQ5Zato4Or}{online} \cite{HWD+}. Along with the data set and a simple \textit{Read me}, the repository also contains a datasheet built according to the recommendation of Gebru \& al. \cite{gebru2021}. This datasheet ensures that the information the researchers need to make informed decisions about how they use the data set in transparently available. It answers most questions a researcher could have about the motivation, the composition of the data, the collection process, the preprocessing, its possible uses, where it is distributed and how it is maintained.

\subsection{Data gathering}

Our data gathering efforts were drastically affected by the 2020 COVID-19 pandemic. We hoped to sample a large number of volunteers, had already bought the necessary material and planned our data gathering procedures. Unfortunately, the social distancing efforts forced us to settle on a smaller size data set with a reduced number of writers that were not randomly sampled. For this reason, it would not be reasonable to use this data set to establish causality. Thankfully, we can still establish the predictability of various variables, compare computer vision models and much more.  

\bigskip

\begin{figure}[ht]
\begin{subfigure}{.5\textwidth}
\centering
\includegraphics[scale=0.3]{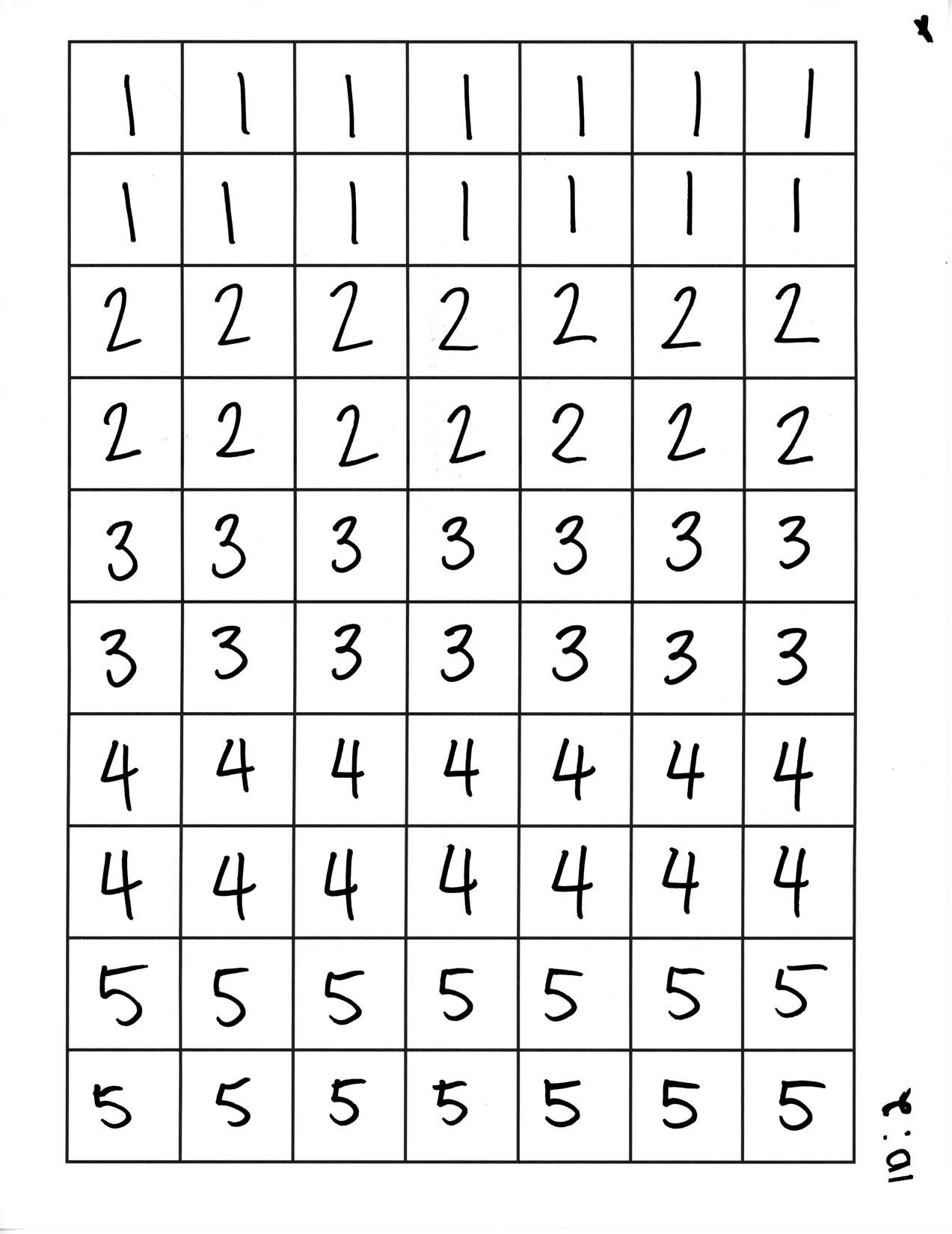}
\end{subfigure}%
\hspace{1em}
\begin{subfigure}{.5\textwidth}
\centering
\includegraphics[scale=0.3]{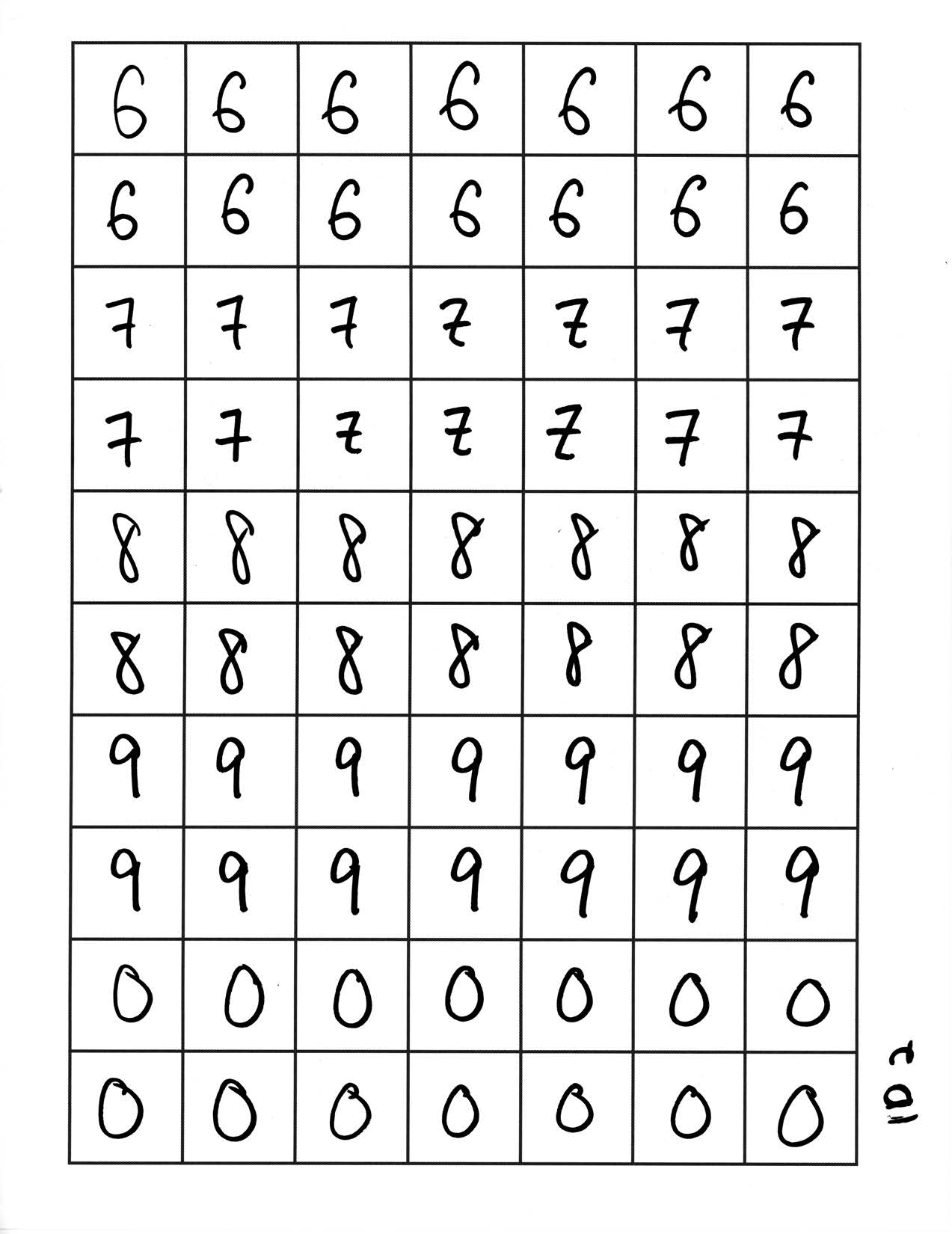}
\end{subfigure}
\caption{Example of the collected images for a single writer.\label{data}}
\end{figure}

Outside of uncontrollable events we made sure to gather a data set in a standardized manner that we believe contains interesting information. Every writer was given 2 pages containing a one inch square grid of 10 rows by 7 columns. Writers were asked to fill these pages with digits, 2 rows per digits for a total of 14 replications per digits as seen in Figure \ref{data}. Every writer was given a new Sharpie pen. 

 We believe the natural handwriting variability for a single task is quite small and subtle and to capture it we need to fix as many external factors as possible. The methodology helps alleviate the variability caused by the writers' mood, pen, paper, the assigned space and other external factors. It provides an environment where the left over variability is closer to the between-replication natural variability than the variability caused by different external factors. 

\bigskip

Following this, the pages of handwritten digits were attached to their user identification (ID). We also collected the following writers characteristics : (1) age, (2) biological gender, (3) height, (4) language taught in elementary school, (5) handedness of the writer, (6) education level and (7) main medium used to write. 

\bigskip

Characteristics (1), (2), (3)  are self-explanatory. Even though we assume that those are not correlated with the handwriting style we collected those variables regardless and we briefly investigate that assumption in Section \ref{Sec:Exp}. For characteristic (4) we were interested to find out if different educational system led to different digit writing styles. Handedness, characteristic (5), was obviously included. Once again we assume the handedness might affect handwritten style and we want to investigate that assumption. The educational level (6) was encoded as a four-level categorical variables were the first level represents high school, the second level means the writer completed an undergraduate program, the third level is assigned to writers who completed a master’s degree or a Ph.D and we finally added a fourth level for young kids who did not complete high school yet. Though the signal might be weak for that variable, we may see some signal coming from young kids, which could also be identified using characteristic (1). Finally, for the most commonly used writing medium (7), writers were asked to choose between handwriting, keyboard or other where participants selecting the latest category commonly reported cellphone or electronic pen. 

\bigskip

 We plan to update our database when the COVID mandates become less restrictive in order to further increase data size. Currently the data set available contains 97 different writers for a total of 13,580 images.  

\subsection{Data processing}

All of the pages collected were scanned using the same machine with the same settings: shades-of-gray and 600 pixels per inches. These pages were then processed through a script that would take off the edges of the pages and divide the grid into 600 by 600 pixels squares. We trimmed of 50 pixels off the four sides of every image to trim off the actual grid and the result is a collection of 500 by 500 pixels images. 

\bigskip

Those images were imported in Python where they were attached to their writer ID, the seven characteristics previously discussed and the digit label. These images are stored as shades of grey images, thus they are composed of a single channel taking values between 0 and 255. When images are scanned, some of the white parts of the images lose some of their purity and thus we have set to 255 every pixel that had a value above 200. The digits were not centred, not scaled and not rotated. These 500 by 500 pixels images form the complete data set available. 

\bigskip

For simplicity we also produced two other data sets with different images size. One data set contains 100 by 100 pixels images. This still is a rather high resolution but it is much faster to run computer vision algorithms on these images than on their 500 by 500 counterparts. We also produced a data set of size 28 by 28 as it is the size of images in the MNIST data set. This allows researchers to use already existing code set up for MNIST and simply swap data sets. The 28 by 28 data set could also be used in conjecture with the MNIST data set for semi-supervised projects. The fact that it is similar to MNIST but very different at the same time should allow us to understand the problems related to the massive use of MNIST in the recent years. Image compression was done using the open CV \cite{opencv} Python library. 

\bigskip

We have done very few pre-processing compared to other popular data sets by choice. To begin, we believe that size and skwedness are genuine writing characteristics that might contain valuable information about the writer and we did not want to discard that information. Thus, we decided to release the data sets detailed above with as little pre-processing as possible.  

\bigskip

\begin{figure}
\centering
\includegraphics[scale=0.8]{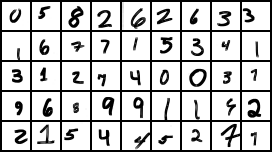}
\caption{Sample of forty-five 28 by 28 pixels images.\label{sample}}
\end{figure}

Figure \ref{sample} contains a sample of what the images in the data set look like.

\subsection{Descriptive statistics}

To conclude our introduction of the HDW+ data set, we provide some simple descriptive statistics. 

\bigskip

\begin{table}[ht]
\centering
\begin{tabular}{|c|c|c|c|c|}
  \hline
    \multirow{2}{*}{Biological Gender} & Male  & Female & & \\
    & 46 (47.4\%) & 51 (52.6\%) & & \\
     \hline
     \multirow{2}{*}{Handiness} & Right & Left & & \\
    & 84 (86.6\%) & 13 (13.4\%) & & \\
         \hline
        \multirow{2}{*}{Language (education)} & French & English &  Other & \\
    & 75 (77.3\%) & 16 (16.5\%) & 6 (6.2\%) & \\
         \hline
        \multirow{2}{*}{Education Level} & No high school  & High school & Bachelor & Graduate \\
    & 7 (7.2\%) & 13 (13.4\%) & 55 (56.7\%) & 22 (22.7\%) \\
         \hline
        \multirow{2}{*}{Usual writing medium} & Hand  & Keyboard & Other & \\
    & 44 (45.4\%) & 45 (46.4\%) & 8 (8.2\%) & \\
         \hline
   
\end{tabular}
\caption{Table of occurrence (proportion) at the time of submission. \label{occ}}
\end{table}

\begin{figure}
\begin{subfigure}{.5\textwidth}
  \centering
  \includegraphics[width=.975\linewidth]{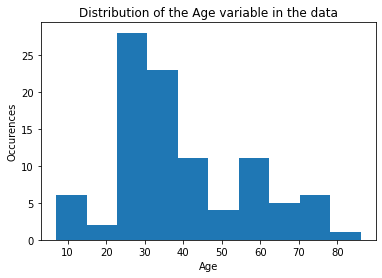}
  \caption{Histogram of the Age variable.}
  \label{fig:sfig1}
\end{subfigure}%
\begin{subfigure}{.5\textwidth}
  \centering
  \includegraphics[width=.975\linewidth]{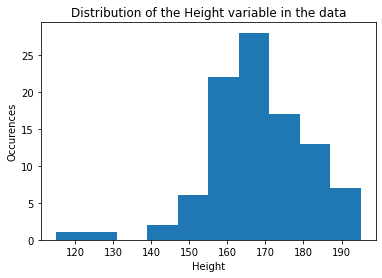}
  \caption{Histogram of the Height variable.}
  \label{fig:sfig2}
\end{subfigure}
\caption{Histograms for continuous variable.}
\label{histos}
\end{figure}

Table \ref{occ} contains the possible values and occurrences of the categorical variables collected and figure \ref{histos} contains the histograms for both continuous variables. Jointly they can provide a quick overview of the distributions of the writer characteristics in our data set. 

\section{Experiments} \label{Sec:Exp}

There are two main purposes for these experiments. First, these experiments act as data exploration: we use them to detect some of the patterns there might exist and establish the first benchmarks for some of the classification problems. Second, in these experiments we showcase some of the new problems and research questions that could not be addressed using the data sets introduced in Section \ref{litrev}.  

\bigskip

All of our experiments were performed using Python \cite{Python} and the Pytorch library \cite{Pytorch}. After experimenting with multiple optimizers, we settled on Adam \cite{kingma14}.

\subsection{Data exploration}

In this section, we explore our data and establish the first benchmarks attainable for various classification tasks. We approach multiple simple classification problems using four models that were previously successful; we implemented Le-Net5 \cite{Lecun98}, a deep fully-connected NN based on the work of Ciresan et al. \cite{Ciresan10}, a committee of 25 CNN \cite{Ciresan12} and finally, we included kNN classification. Le-Net5 \cite{Lecun98} was selected as our default CNN; it is introduced in the same paper that introduced the MNIST data set. We included a deep NN based on the work of Ciresan et al. \cite{Ciresan10} who demonstrated that a very deep and large NN performs as well as a CNN for digit prediction.  We included a committee of CNN since ensemble models have the best classification accuracy on the MNIST data set. Finally, the kNN classifier was added to serve as a simple benchmark to quickly see how simple some of the prediction tasks are.

\bigskip

Second we address some possible new problems we can approach with this new data set. We assess how higher resolution affects classification performances and how using writer characteristics as predictors affects the prediction accuracy. We do not address multi-label classification problems in this article, but this is another problem that can be tackled with this data set that could not be tackled with MNIST. 

\bigskip

The single Le-Net5 CNN, the deep NN and the kNN classifier are fit 50 times where each time we randomized which images are in the training set and the testing set. We fit the ensemble model 15 times with once again randomized training and test set for each trial. Repeated random subsampling allows us to adjust the size of the training set and test sets independently from the number of Monte Carlo samples. In contrast to k-fold cross-validation, this allows us to use large training and testing sets without having only few Monte Carlo samples.

\subsubsection{Image classification}\label{clas}

As already mentioned we wish to establish the first benchmark but also the existence of some signal; thus we often time compare our results with the \textit{naive classifier}, which we define here as a classifier that always votes on the majority class. Readers can look at the descriptive statistics of Table \ref{occ} to get a rough idea of the performance of such naive classifier in this analysis.

\bigskip

For some of these experiments, we divide our data set into a training set and a testing set in a way that both sets contain every writer; the training set contains 10 images of every digit of every writer and the test set contains 4 images of every digit of every writer. We named this process \textit{partitioning by digits}. This partitioning will be used when predicting the ID. To better assess the actual predictability of the writer characteristics we created another way to partition training data from test data; this time we split training and test sets by participants, randomly assigning 70\% of the writers to be in the training set and 30\% in the test set. This way the writers in the test set have never been observed during training. We refer to this as \textit{partitioning by individuals}.

\begin{table}[ht]
\centering
\begin{tabular}{|c|cc|cc|cc|cc|}
  \hline
   \multirow{2}{1em}{}  & \multicolumn{2}{c|}{kNN }  & \multicolumn{2}{c|}{LeNet-5} & \multicolumn{2}{c|}{Comm. LeNet-5} & \multicolumn{2}{c|}{Deep NN}   \\ 
    & Mean & Std & Mean & Std & Mean & Std  & Mean & Std  \\ 
  \hline
    Digit & 0.7431 & 0.0222 & 0.9399  & 0.0143 & 0.9762 & 0.0013 & 0.9192 & 0.0160  \\
    \hline
    ID & 0.1682 & 0.0049 & 0.3473 & 0.0136 & 0.6195 & 0.0063& 0.4268 & 0.0077  \\
    \hline
    Gender & 0.5608 & 0.0124 & 0.5367 & 0.0183 & 0.5483 & 0.0372 & 0.5394 & 0.0208    \\
    \hline
    Language & 0.6971 & 0.0369 & 0.6792 & 0.0322 & 0.7621 & 0.0626  & 0.6752 & 0.0604   \\
    \hline
    Hand & 0.7004 & 0.0131 & 0.7940 & 0.0285 & 0.8304  & 0.0499  & 0.7973 & 0.0275   \\
    \hline 
    Education Level & 0.4265 & 0.0347 & 0.4117 & 0.0222 & 0.4726 & 0.0343  & 0.4147 & 0.0393  \\
    \hline
    Writing Medium & 0.4650 & 0.0243 & 0.4585 & 0.0225 & 0.4782 & 0.0372 & 0.4668 & 0.0189  \\
    \hline
\end{tabular}
\caption{Mean and standard deviation of the prediction accuracy for simple classification tasks. \label{classpred}}
\end{table}

\bigskip

Let us start with kNN, which serves as an exploratory data analysis approach. When kNN achieves a good performance compared to the naive classifier, it demonstrates that the problem is relatively simple or at least that there exist some signal that can possibly be investigated further. The kNN model performs well compared to the naive classifier on digit and ID classification, indicating those variables can be predicted. However, kNN performs similarly to every model, including the naive classifier, when predicting gender and writing medium. This is a sign we cannot predict those variables, or that the signal is quite small compared to the noise. 

\bigskip

Overall, in the table \ref{classpred} we see that this data set has a very high signal with respect to the digit. HWD+ is of much smaller size and less processed than MNIST but nonetheless a committee of CNNs reaches close to 98\% accuracy on average and is significantly better than the second-best model (p-value $< 0.001$). 

\bigskip

Next, we look at writer identification. The performance of the committee is quite impressive as observed in table \ref{classpred}, accurately predicting the writer 62\% of the time, given we have a pool of 97 writers this is way above the performances of the naive classifier. The committee of LeNet-5 is also significantly better than the second-best model (p-value $< 0.001$). 

\bigskip

Let us now discuss the prediction of the various characteristics collected. As we previously discussed, we used a different data set partitioning for the writer characteristics. The reason is quite simple; the CNN techniques are so good at predicting distinct style related to IDs, as shown by the high performance of the committee, that the model could map images to IDs and then IDs to characteristics. This is not exactly identifying writing patterns that are specific to some of the writer characteristics. Thus, we implemented \textit{partitioning by individuals} for writer characteristics to make sure the algorithm actually tries to learn the effect of the characteristics on the writing styles that are shared among writers. 

\bigskip

Most of the results are even or worse than the naive classifier who simply selects the majority class. However, we noticed in table \ref{classpred} that the improvement when using a committee of CNNs over a single CNN is statistically significant when predicting every characteristic except gender and writing medium; thus there might some signal for native language (p-value $< 0.001$), handedness (p-value = 0.008) and education level (p-value $< 0.001$). We believe that this significant improvement of the committee model over the second best is a consequence of the existence of a signal. Thus, even though currently none of the models outperform the naive classifier, there might be some ways to further improve the prediction accuracy in order to surpass it for those classification tasks. 

\bigskip

Based on this simple exploration, our proposed data set contains  variables with various degrees of predictability: we can achieve high accuracy when predicting the digit, the ID seems to lead to widely different prediction performances but is predictable, some characteristics, such as native language, handedness and education level, are weakly related to the images and finally the gender and usual writing medium seem to be difficult to predict using only images of digits. 

\bigskip

We also noticed the relatively good performances of the deep NN which supports the results of Ciseran et al. \cite{Ciresan10}. This shows that our data set can be used to assess if some results observed on MNIST can be generalized to other handwritten digits data set. 

\subsubsection{High resolution images classification}

In the next two sections, we showcase tasks that are specific to our new data set. In this section we explore the effect of higher resolution images on the classification performances of CNNs. Being able to provide users with images as high resolution as 500 by 500 pixels is something offered by very few data sets that often contain very small images. However, in this section what we call high-resolution images are 100 by 100 pixels images. 

\bigskip

\begin{table}[ht]
\centering
\begin{tabular}{|c|cc|cc|cc|}
  \hline
   \multirow{2}{1em}{} & \multicolumn{2}{c|}{LeNet-5 (28x28)} & \multicolumn{2}{c|}{Comm. (28x28)} & \multicolumn{2}{c|}{LeNet-5 (100x100)}   \\
    & Mean & Std & Mean & Std & Mean & Std  \\ 
  \hline
    Digit & 0.9399  & 0.0143 & 0.9762 & 0.0013 & 0.9683 & 0.0044  \\
    \hline
    ID & 0.3473 & 0.0136 & 0.6195 & 0.0063  & 0.3675 & 0.0224  \\
    \hline
    Gender & 0.5367 & 0.0183 & 0.5483 & 0.0372 & 0.5354 & 0.0410   \\
    \hline
    Language & 0.6792 & 0.0322 & 0.7621 & 0.0626 & 0.7284 & 0.0441   \\
    \hline
    Hand & 0.7940 & 0.0285 & 0.8304  & 0.0499 & 0.8129  & 0.0355  \\
    \hline 
    Education Level & 0.4117 & 0.0222 & 0.4726 & 0.0343 & 0.4466 & 0.0368   \\
    \hline
    Writing Medium & 0.4585 & 0.0225 & 0.4782 & 0.0372 & 0.4612 & 0.0234   \\
    \hline
\end{tabular}
\caption{Mean and standard deviation of the prediction accuracy for simple classification tasks on low-resolution images (single LeNet-5 and committee) compared to high-resolution images (single LeNet-5). \label{HRPred}}
\end{table}

We observe a statistically significant increase in prediction accuracy for the high-resolution image over the low-resolution one when predicting the digit (p-value $< 0.001$), the writer ID (p-value $< 0.001$), the first language (p-value $< 0.001$), the writer handedness (p-value = 0.002) and the education level of the writer (p-value $< 0.001$) in table \ref{HRPred}. These are the exact same variables for which the committee also improved on the benchmark LeNet-5 in table \ref{classpred}. Even though the predictive performance of LeNet-5 is equivalent to the naive classifier for some of those variables, those improvements when using a committee or higher resolution images lead us to believe that those variables are predictable in some way. In other words, there must exist some signal between the images and those variables. 

\bigskip

The results here are intuitive: the classifier benefits from higher resolution images since they are richer in information. However, it should not be surprising that it also increased the computational cost. Even though computational power is ever increasing, so is the size of our data sets and consequently computational scalability will always be a concern. Our data set offers the opportunity to analyse the scalability of data and models on a simple digit prediction task. 

\bigskip

We noticed training a single CNN on the high-resolution images is 25 times slower than training a single CNN on the low-resolution images. Consequently, training a committee of 25 CNNs on the low-resolution images takes a similar amount of time than training a single CNN on high-resolution images. We see in table \ref{HRPred} that the ensemble of classifiers trained on the low-resolution data set performs better than the single LeNet-5 trained on a richer data set. We expect a committee of LeNet-5 trained on the high-resolution data set to have higher performance than the committee training on the low-resolution data set, but this is not the point we are trying to get across. Our results reveal that the predictive improvement provided by using an ensemble technique is higher than the improvement provided by getting a data set with twelve times as many pixels for a fixed run-time. To conclude, we believe this is an interesting concept to study in future work: striking the right balance between image resolution and model complexity considering the fact that limited computing power will always be a concern.

\subsubsection{Image classification with predictors}

In this section we include some of the collected information as predictors to see how it changes the performances of the LeNet-5 classifier, once again something new that our data set enables. Moreover, we are interested in understanding the potential contribution of additional information in images classification. It is common in the prediction of neurodegenerative diseases to combine image data (such as MRI) with patients’ characteristics \cite{el2021}. In this section, we explore this idea of combining image data with ancillary variables. 

\bigskip

We experiment with two simple tasks: in the first experiment we try to classify images according to their digit and we incorporate the writer ID as an additional predictor. Next, we do the opposite, we classify images according to the writer ID while including the digit as an additional predictor. To do so, we include a one-hot encoding vector for writer ID or the digit in the first fully connected layer of LeNet-5, after the convolutional layers. For this experiment, we \textit{partitioned by digits} the data set.  

\bigskip

\begin{table}[ht]
\centering
\begin{tabular}{|c|cc|cc|cc|cc|}
  \hline
   \multirow{2}{*}{} & \multicolumn{2}{c|}{Images (LeNet-5)} & \multicolumn{2}{c|}{Images + (LeNet-5)}  & \multicolumn{2}{c|}{Images (Com.)} & \multicolumn{2}{c|}{Images + (Com.)}  \\
    & Mean & Std & Mean & Std  & Mean & Std & Mean & Std  \\ 
  \hline
    Digit & 0.9399  & 0.0143  & 0.9551 & 0.0080 & 0.9762 & 0.0013 & 0.9812 & 0.0020 \\
    \hline
    ID & 0.3473 & 0.0136  & 0.3575 & 0.0192 & 0.6195 & 0.0063 & 0.6003 & 0.0042 \\
    \hline
\end{tabular}
\caption{Mean and standard deviation of the prediction accuracy for simple classification tasks when using only the image as predictors (Images) or the image and an additional predictor (Images +)\label{predpred}}
\end{table}

\bigskip

Including the writer ID as additional information significantly increases the accuracy when predicting the digit (p-value $< 0.001$ for both models). However, including the digit when predicting the ID actually decreased the prediction accuracy of the committee.

\bigskip

These results warrant further investigation. For instance, there exist multiple way to integrate additional information in a CNN classifier and combining modalities of different types still is a challenge to this day. Our data set offers an opportunity to explore those research questions further.

\subsection{Semi-supervised exploration}

In this section, we explore how to combine our data set with some other handwritten digit data sets, which is a possibility offered by HWD+. Though there are many ways to combine data sets, we will explore a semi-supervised application in that section. We treat our HWD+ as labelled data ($S_l$) and the MNIST data set as unlabelled data ($S_u$). To divide the HDW+ in a training and a testing set, we used the \textit{partitioning by digits} when predicting the ID and \textit{partitioning by individuals} when predicting the digit as described in section \ref{clas}.

\bigskip

In this section, we used the M2 model proposed by Kingma \cite{kingma14,Kingma17}, briefly introduced below:

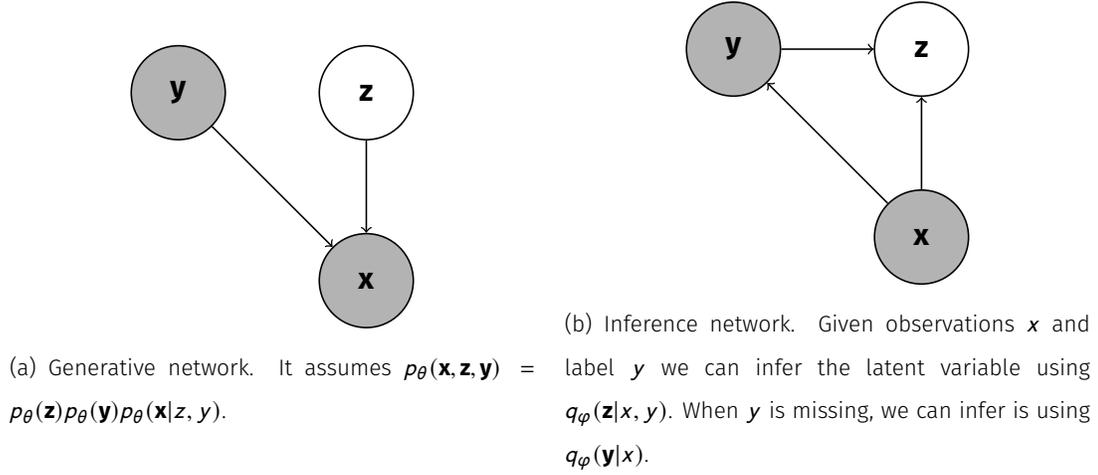
\begin{figure}[H]
\begin{subfigure}{.5\textwidth}
  \centering
  \begin{tikzpicture}[->, semithick]
  \tikzstyle{latent}=[fill=white,draw=black,text=black,style=circle,minimum size=1.25cm]
  \tikzstyle{observed}=[fill=black!30,draw=black,text=black,style=circle,minimum size=1.25cm]

  \node[latent]   (A) at (0,2.5)  {\large $\mathbf{z}$};
  
    \node[observed]         (B) at (-2.5,2.5)  {\large $\mathbf{y}$};

  \node[observed]         (C) at (0,0)  {\large $\mathbf{x}$};

  \path (A) edge            node {} (C);
  \path (B) edge            node {} (C);

\end{tikzpicture}
  \caption{Generative network. It assumes $p_\theta(\mathbf{x},\mathbf{z},\mathbf{y}) = p_\theta(\mathbf{z})p_\theta(\mathbf{y})p_\theta(\mathbf{x}|z,y)$.}
  \label{genny}
\end{subfigure}%
\hspace{1em}
\begin{subfigure}{.5\textwidth}
  \centering
  \begin{tikzpicture}[->, semithick]
  \tikzstyle{latent}=[fill=white,draw=black,text=black,style=circle,minimum size=1.25cm]
  \tikzstyle{observed}=[fill=black!30,draw=black,text=black,style=circle,minimum size=1.25cm]

  \node[latent]   (A) at (0,2.5)  {\large $\mathbf{z}$};
  
    \node[observed]         (B) at (-2.5,2.5)  {\large $\mathbf{y}$};

  \node[observed]         (C) at (0,0)  {\large $\mathbf{x}$};

     \path (C) edge            node {} (A);
  \path (C) edge            node {} (B);
  \path (B) edge            node {} (A);     

\end{tikzpicture}
  \caption{Inference network. Given observations $x$ and label $y$ we can infer the latent variable using $q_\varphi(\mathbf{z}|x,y)$. When $y$ is missing, we can infer is using $q_\varphi(\mathbf{y}|x)$.}
  \label{inferency}
\end{subfigure}
\caption{Graphical representation of the two networks that makes up the M2 model.\label{M2}}
\end{figure}

The M2 model is trained using the following objective function:

\begin{equation} 
\mathcal{J}^\alpha =  \sum_{S_l} \mathcal{L}(x,y) + \sum_{S_u} \mathcal{U}(x) + \alpha\mathbf{E}_{S_l}\left[-\ln q_\varphi(\mathbf{y}|x)\right],
\label{m2eq}
\end{equation}

where $\alpha$ is a hyper-parameter that controls the relative weight between generative and discriminative learning, where $\mathcal{L}(x,y)$ is the objective function on labelled observations $S_l$:

\begin{equation}
\mathcal{L}(x,y) = \mathbf{E}_{q(\mathbf{z}|\mathbf{x},\mathbf{y})}\left[ \ln p_\theta(\mathbf{z})+ \ln p_\theta(\mathbf{y}) + \ln p_\theta(\mathbf{x}|z,y) - \ln q_\varphi(\mathbf{z}|x,y)\right],  \\
\end{equation}

and where $\mathcal{U}(x)$ is the objective function for unlabelled observations $S_u$:

\begin{equation}
\begin{split}
\mathcal{U}(x) &=  \mathbf{E}_{q(\mathbf{z},\mathbf{y}|x)}\left[ \ln p_\theta(\mathbf{z})+ \ln p_\theta(\mathbf{y}) + \ln p_\theta(\mathbf{x}|z,y) - \ln q_\varphi(\mathbf{z},\mathbf{y}|x)\right] \\
&= \sum_y \left[q_\varphi(\mathbf{y}|x)(\mathcal{L}(x,y))\right] + \mathcal{H}(q_\varphi(\mathbf{y}|x)), 
\end{split}
\end{equation}
where $\mathcal{H}$ is the entropy of the distribution.

\bigskip

The bigger $\alpha$ is in equation \ref{m2eq} the closer we are to obtain the same classifier obtained using strictly the labelled data; in a way the whole VAE machinery can be perceived as regularization that prevents overfitting the training labelled point. More details about the M2 model can be found in various publications \cite{kingma14,Kingma17,rastgoufard18}.

\bigskip

We use the M2 model  to predict the digit and the ID in our images while increasing our data set size with some unlabelled images, the MNIST data set. In our implementation of the M2 model $q_\varphi(\mathbf{y}|x)$ is parametrized by a LeNet-5 CNN. We assess the improvement produced when including new unlabelled data compared to the results previously obtained in Section \ref{clas} when using a single LeNet5. 

\bigskip

This gives us a great perspective on semi-supervised classification. It is said that it is possible to leverage unlabelled points from other data sets to improve the accuracy over the simple classifier and we have argued it is due to some regularization. However fitting the compression and decompression machinery does increase the run time needed to fit such semi-supervised model.

\bigskip

\begin{table}[ht]
\centering
\begin{tabular}{|c|cc|cc|}
  \hline
   \multirow{2}{1em}{} & \multicolumn{2}{c|}{LeNet-5} & \multicolumn{2}{c|}{M2}  \\
    & Mean & Std & Mean & Std  \\ 
  \hline
    Digit & 0.9399  & 0.0143 & 0.9542 & 0.0060 \\
    \hline
    ID & 0.3473 & 0.0136 & 0.4174 & 0.0099  \\
    \hline
\end{tabular}
\caption{Mean and standard deviation of the prediction accuracy of the semi-supervised M2 model trained on the HWD+ and MNIST data set compared to LeNet-5 trained on HWD+. \label{SSPred}}
\end{table}

Table \ref{SSPred} shows a significant increase in accuracy when using the semi-supervised model (p-value $< 0.001$ in both cases). These results are surprising for us given how different the MNIST data set is from our data set with respect to many aspects. The second term of the objective function presented in equation \ref{m2eq} trains the classifier on labelled data and is precisely what we trained in previous sections. Further investigation on how the first term serves as regularization should lead to interesting results. Our data set makes possible the study of the behaviour of semi-supervised models in such applications on different prediction tasks.  

\subsection{Conditional image generation}

In this section we showcase the opportunity our data set offers for controllable (conditional) image generation. We briefly discuss and demonstrate the generative abilities of the SGDM model using our data set. The multiple labels allow us to turn multiple \textit{control knobs} which imbue the generative process with much more control, consequently this data set is a contribution towards conditional image generation research. 

\bigskip

For this experiment, we implemented the SDGM proposed by Maal{\o}e et al. \cite{maaloe15,maaloe16,rastgoufard18}. The model relies on auxiliary variables \cite{agakov04} to improve the expressive power of both the inference and generative model. 

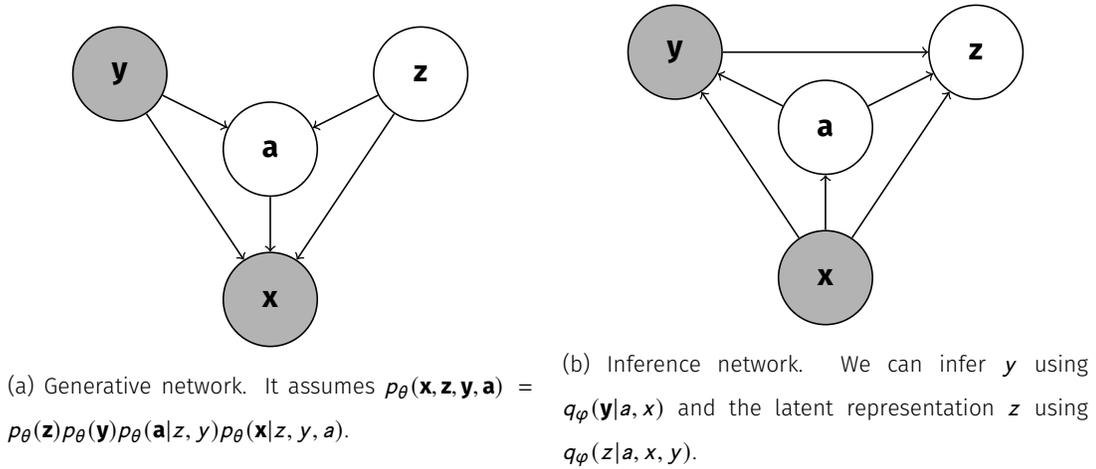
\begin{figure}[H]
\begin{subfigure}{.5\textwidth}
  \centering
  \begin{tikzpicture}[->, semithick]
  \tikzstyle{latent}=[fill=white,draw=black,text=black,style=circle,minimum size=1.25cm]
  \tikzstyle{observed}=[fill=black!30,draw=black,text=black,style=circle,minimum size=1.25cm]

  \node[latent]   (A) at (2,3)  {\large $\mathbf{z}$};
  
    \node[observed]         (B) at (-2,3)  {\large $\mathbf{y}$};

  \node[observed]         (C) at (0,0)  {\large $\mathbf{x}$};
  
    \node[latent]          (D) at (0,2)  {\large $\mathbf{a}$};

  \path (A) edge            node {} (D);
  \path (A) edge            node {} (C);
  \path (B) edge            node {} (D); 
  \path (B) edge            node {} (C); 
  \path (D) edge            node {} (C);

\end{tikzpicture}
  \caption{Generative network. It assumes $p_\theta(\mathbf{x},\mathbf{z},\mathbf{y},\mathbf{a}) = p_\theta(\mathbf{z})p_\theta(\mathbf{y})p_\theta(\mathbf{a}|z,y)p_\theta(\mathbf{x}|z,y,a)$.}
  \label{genny}
\end{subfigure}%
\hspace{1em}
\begin{subfigure}{.5\textwidth}
  \centering
  \begin{tikzpicture}[->, semithick]
  \tikzstyle{latent}=[fill=white,draw=black,text=black,style=circle,minimum size=1.25cm]
  \tikzstyle{observed}=[fill=black!30,draw=black,text=black,style=circle,minimum size=1.25cm]

   \node[latent]   (A) at (2,3)  {\large $\mathbf{z}$};
  
    \node[observed]         (B) at (-2,3)  {\large $\mathbf{y}$};

  \node[observed]         (C) at (0,0)  {\large $\mathbf{x}$};
  
    \node[latent]          (D) at (0,2)  {\large $\mathbf{a}$};

  \path (C) edge            node {} (D);
  \path (C) edge            node {} (B);
  \path (C) edge            node {} (A); 
  \path (D) edge            node {} (A); 
  \path (D) edge            node {} (B); 
  \path (B) edge            node {} (A); 
\end{tikzpicture}
  \caption{Inference network. We can infer $y$ using $q_\varphi(\mathbf{y}|a,x)$ and the latent representation $z$ using  $q_\varphi(z|a,x,y)$.}
  \label{inferency}
\end{subfigure}
\caption{Graphical representation of the two networks that makes up the SDGM model.\label{SDGM}}
\end{figure}

Figure \ref{SDGM} is a graphical representation of the SDGM. Just like the M2 model, the objective function has a component for labelled observations, a component for unlabelled observations and an extra term to ensure that $q_\varphi(\mathbf{y}|a,x)$ is trained with labelled observations. More details about the SDGM can be found in various publications \cite{maaloe15,maaloe16,rastgoufard18}.

\bigskip

 We fit the SDGM with both the ID and the digit as labels $\mathbf{y}$. Since the model is fitted for generative purpose, we use all of our data points, which are labelled, and the classifier $q_\varphi(\mathbf{y}|x)$ is completely irrelevant here. What we truly want is to train $p_\theta(\mathbf{x}|z,a,y)$ to generate images that are good looking and that respect the conditions imposed by $y$. In other words, the images have to be of the right digit with the right style. Other details of the images are randomized through $\mathbf{z}$ and $\mathbf{a}$. 

\bigskip

To showcase our results we have produced the figures below. We picked four different IDs with drastically different styles to better illustrate that the algorithm was able to grasp some writing style details. In the figures below, the first four columns are a sample of four real images and the six following columns are generated images. We have selected the digits one, two, four, seven and nine has they exhibit large differences in style from one writer to another. 

\begin{figure}[H]
\centering
\includegraphics[height=5cm]{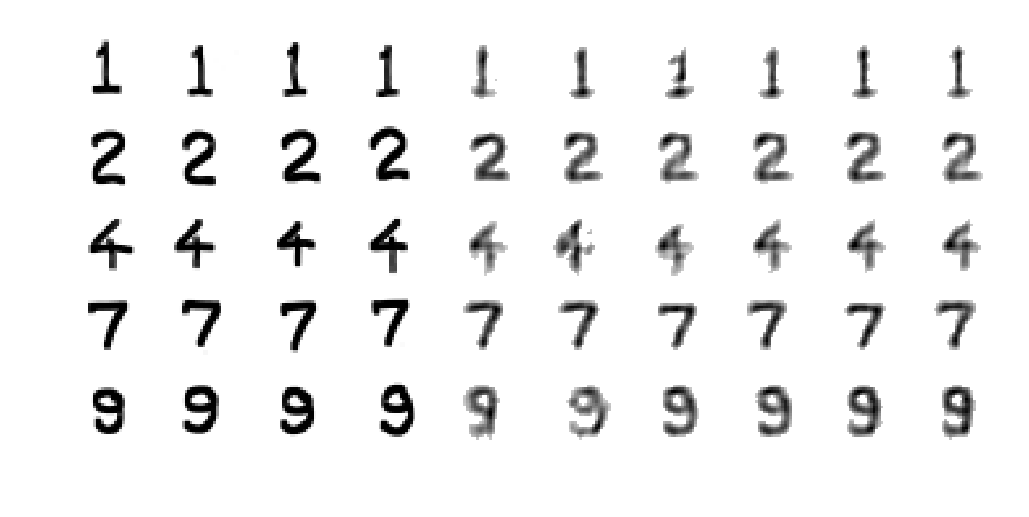}
\caption{Generated images for ID \#12 \label{Gen12}}
\end{figure}

\begin{figure}[H]
\centering
\includegraphics[height=5cm]{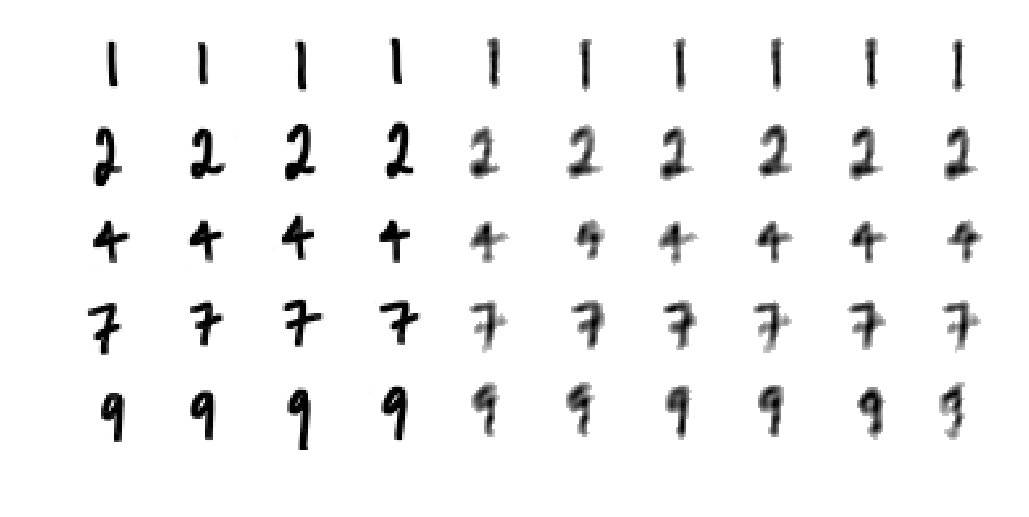}
\caption{Generated images for ID \#14 \label{Gen14}}
\end{figure}

\begin{figure}[H]
\centering
\includegraphics[height=5cm]{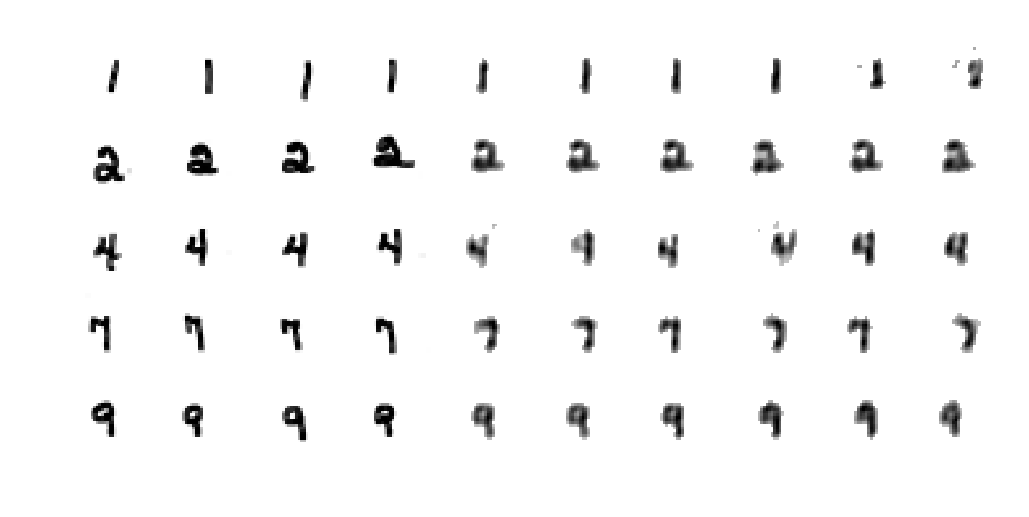}
\caption{Generated images for ID \#29 \label{Gen29}}
\end{figure}

\begin{figure}[H]
\centering
\includegraphics[height=5cm]{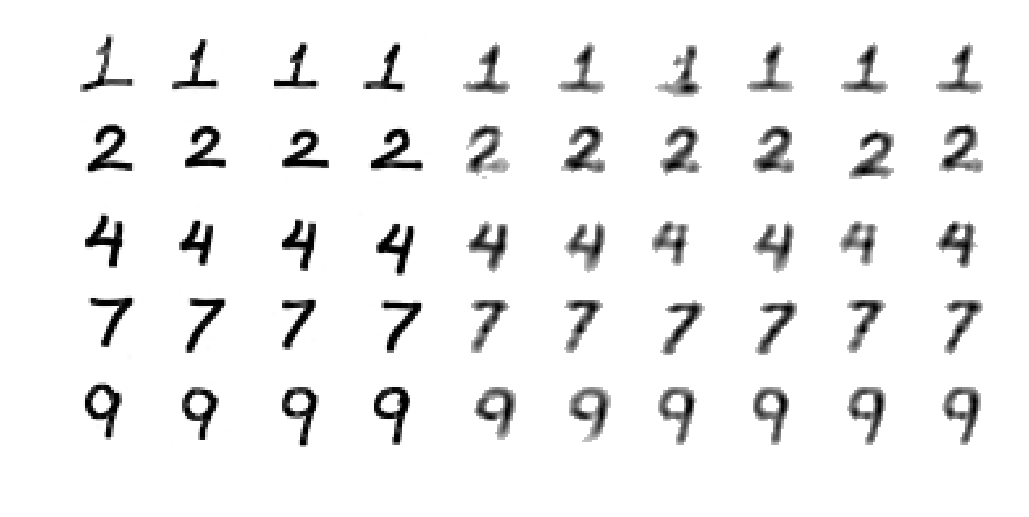}
\caption{Generated images for ID \#70 \label{Gen70}}
\end{figure}

The generator seemed to have learned very well the effect of the digit input. We see that the generated digits are distinguishable and appropriate. This was to be expected based on previous experiments \cite{Kingma17}.

\bigskip

Additionally, the SDGM also learned the writing styles of the various writers as observed in Figures \ref{Gen12},\ref{Gen14},\ref{Gen29} and \ref{Gen70}. We observe that the size of generate images respect the size of the true images as well as multiple details such as serifs and angles. For instance, the images of \textit{ones} generated by the SDGM model has serif for ID \#12 and ID \#70 and not the other two. Similarly the \textit{fours} are open for ID \#29 and \#70 but closed for ID \#12 and \#14. Moreover, \textit{sevens} take all kinds of shape, sometimes the tail of the digit \textit{nine} is curved and so fort. Overall we are pleased with the results. We already knew it was possible to generate images of a specified digit but the writer ID is something more subtle and those images prove that the VAE model is able to grasp and mimic what makes writing styles different. 

\bigskip

However, the generated images are blurry but this is a well-known problem for VAE generated images \cite{sohn15,huang18,dorta18,dorta18b} and a problem we are not trying to fix in this article. The generative process could be further improved with new upcoming VAE structure \cite{dorta18} or other generative models such as GANs \cite{goodfellow14} which do not suffer as much from the blurry images problem. 

\bigskip

The purpose of this experiment is not to produce the best-looking images but rather to assess how conditional image generation can be tested on our data set. Our results are preliminary but they highlight the capacity of some well-developed generative models to grasp subtle writing styles and the opportunity that our data provide to experiment with such generative models. Overall, we believe our data set is a good playground for conditional image generation. 

\section{Conclusion}

In this article we introduced a brand new data set, HWD+, which contains almost 14 000 high-resolution images of handwritten digits attached to a set of labels containing the digit, the writer ID and various writer characteristics. The data set has been carefully collected and processed and is publicly available online.

\bigskip

We have done a first analysis of the data set; we showed that our data contains variables with different predictability making it a useful alternative to MNIST for testing new computer vision algorithms. We especially considered classification tasks that were made possible with our new data set such as including additional predictors in classification tasks or using higher-resolution images. 

\bigskip

We have also proceeded with a semi-supervised analysis. We have shown the potential use of our data set in a semi-supervised classification task in tandem with the MNIST data set; the use of the M2 model led to a more accurate LeNet-5 classifier. We have also shown the potential of our multi-label data set for conditional image generation. We believe our data set is the perfect testing ground for new creative controllable generative models.

\pagebreak

\section*{Acknowledgement}

The authors gratefully acknowledge the financial support from the Natural Sciences and Engineering Research Council (NSERC) of Canada and the Ontario Student Assistance Program (OSAP). The authors would also like to recognize the contribution of the 150 participants who returned data sheets; without them this new data base wouldn't exist.

\bibliographystyle{plain}

\bibliography{mybibfile}

\end{document}